# An Effective Fingerprint Classification and Search Method

**M H Bhuyan　and D K Bhattacharyya**

Dept of Computer Sc. & Engg., Tezpur University, Napaam, Tezpur, Assam, India

**Summary**
This paper presents an effective fingerprint classification method designed based on a hierarchical agglomerative clustering technique. The performance of the technique was evaluated in terms of several real-life datasets and a significant improvement in reducing the misclassification error has been noticed. This paper also presents a query based faster fingerprint search method over the clustered fingerprint databases. The retrieval accuracy of the search method has been found effective in light of several real-life databases.

## 1. Introduction

Fingerprints have been used as one of the most popular biometric authentication and verification measures because of their high acceptability, immutability and uniqueness [1]. Immutability refers to the persistence of the fingerprints over time whereas uniqueness is related to the individuality of ridge details across the whole fingerprint image. Fingerprint classification is an important step in any fingerprint identification system because it significantly reduces the time taken in identification of fingerprints especially where the accuracy and speed are critical. To reduce the search and space complexity, a systematic partitioning of the database into different classes is highly essential.
Key to the task of classification is the feature extraction. The effectiveness of feature extraction depends on the quality of the images, representation of the image data, the image processing models, and the evaluation of the extracted features. At the first stage of the fingerprint classification process, the image is only represented as a matrix of grey scale intensity values. Feature extraction is a process through which geometric primitives within images are isolated in order to describe the image structure, i.e. to extract important image information and to suppress redundant information that is not useful for classification and identification processes. Thus fingerprint features and their relationships provide a symbolic description of a fingerprint image.
In this paper, we present an effective method to classify the fingerprint images using the hierarchical agglomerative clustering technique and also presents a query based search method over clustered databases. For classification of the fingerprint images, it initially determines a compact representation (a 32-dimensional numeric sequence or pattern) for each fingerprint image based on the ridge flow patterns by exploiting the *freeman chain code* approach [2]. Then it applies enhanced DROCK [3], an agglomerative clustering algorithm to classify the fingerprint images into six classes. The simplified DROCK logic makes the classification method more attractive in view of the following points: (*i*) linear cluster formation time, and (*ii*) smaller clusters generated, reduces the search complexity. Next, we report a brief review on the existing fingerprint classification and searching techniques.

## 2. Related Works

Based on our survey, it has been observed that most of the existing works are aimed to classify the fingerprint database based on the minutiae sets and singular points [4, 5]. In this section, we report some of these in brief.

*[a] Masayoshi Kamijo's approach* [6]: It is an ANN based approach, where a neural network for the classification of fingerprint images is constructed, which has been claimed to be capable of classifying the complicated fingerprint images. It uses a two-step learning method to train the four-layered neural network which has one sub-network for each category of fingerprint images. It carries out the principal component analysis (PCA) with respect to the unit values of the second hidden layer and also studies the fingerprint classification state represented by the internal state of the network. Consequently, the method confirms that the fingerprint patterns are roughly classified into each category in the second hidden layer and also it measures and conforms the effectiveness of the two-step learning process. However, in case of larger datasets, this method can not be found effective.
*[b] Karu and Jain's approach* [7]: This approach first finds the ridge direction at each pixel of an input fingerprint image. Then the algorithm extracts global features such as singular points (*cores* and *deltas*) in the fingerprint image and performs the classification based on the number and locations of the detected singular points. Here, the singular point(s) detection is an iterative regularization process until the valid singular points are detected. If the images are of poor quality, the algorithm classifies those images as unknown types based on some

threshold values. However, the algorithm can detect the labeled images with high quality.

[*c*] *Ballan and Ayhan Sakarya's technique* [8]: Here, a fast, automated, feature-based technique for classifying fingerprints is presented. The technique extracts the singular points (*deltas* and *cores*) in the fingerprints based on the directional histograms. It finds the directional images by checking the orientations of individual pixels, computes directional histograms using overlapping blocks in the directional image, and classifies the fingerprint into the Wirbel classes (*whorl* and *twin loop*) or the Lasso classes (*arch*, *tented arch*, *right loop*, or *left loop*). The complexity of the technique is the order of the number of pixels in the fingerprint image. However, it takes much time for classification.

[*d*] *Jain, Prabhakar and Hong's Multi-channel approach* [9]: This method can be found to be more accurate while classifying the fingerprint images as compared to its previous counterparts. Here, the fingerprints are classified into five categories: *whorl*, *right loop*, *left loop*, *arch*, and *tented arch*. The algorithm uses a novel representation (FingerCode) and is based on a two-stage classifier to make the classification effective. The two-stage classifier uses a *k*-nearest neighbor classifier in its first stage and a set of neural network classifiers in its second stage to classify a feature vector into one of the five fingerprint classes. This algorithm suffers from the requirement that the region of interest be correctly located, requiring the accurate detection of center point in the fingerprint image. Otherwise, the algorithm can be found to be very effective.

[*e*] *Cho, Kim, Bae et al's approach* [10]: They described an effective fingerprint classification algorithm that uses only the information related to the core points. The algorithm detects core point(s) candidates roughly from the directional image and analyzes the near area of each core(s) candidate. In this core analysis, false core point(s) made by noise are eliminated and the type and the orientation of the core point(s) are extracted for the classification step. Using this information, classification was performed. However, it can be found to be very difficult to eliminate the false singular point(s) which has been used for class decision. It demands for more sophisticated methods to eliminate those false core point(s) towards a noise-tolerant classification system.

[*f*] *Yao, Marcialis, Pontil, et al's approach* [11]: Here, a new fingerprint classification algorithm is reported based on two machine learning approaches: support vector machines (SVMs), and recursive neural networks (RNNs). RNNs are trained on a structured representation of the fingerprint image. They are also used to extract a set of distributed features which can be integrated in the SVMs. SVMs are combined with a new error correcting coding scheme which, unlike previous systems, can also exploit information contained in ambiguous fingerprint images.

[*g*] *Tan, Bhanu and Lin's approach* [12]: Here, a fingerprint classification approach was proposed based on a novel feature-learning algorithm. They used Genetic Programming (GP) based approach, which learns to discover composite operators and features that are evolved from combinations of primitive image processing operations. They developed an approach to learn the composite operators based on primitive features automatically. It can be useful in extracting some useful unconventional features, which are beyond the imagination of humans. They also defined the primitive operators as very fundamental and easy to compute. Then, primitive operators are separated from computation at operators and feature generation operators. Features are computed using feature generation operators. However, this classification method can be found to be effective over quality fingerprint images.

[*h*] *Shah and Sastry's approach* [13]: It classifies the fingerprint images into one of the five classes: *arch*, *tented arch*, *left loop*, *right loop*, and *whorl*. It can be found to be useful over low-dimensional feature vector obtained from the output of a feedback based line detector. Their line detector was a co-operative dynamic system that gives oriented lines and preserves multiple orientations at points where differently oriented lines meet. Also the feature extraction process of their method was based on characterizing the distribution of orientations around the fingerprint. Three types of classifiers were used here: support vector machines, nearest-neighbor classifier, and neural network classifier. The line detector works on binary images only.

[*i*] *Park and Park's approach* [14]: Here, a new approach for fingerprint classification is reported based on Discrete Fourier Transform (DFT) and nonlinear discriminant analysis. Utilizing the DFT and directional filters, a reliable and efficient directional image is constructed from each fingerprint image, and then nonlinear discriminant analysis is applied to the constructed directional images, reducing the dimension drastically and extracting the discriminant features. This method explores the DFT and directional filtering in dealing with low-quality images.

[*j*] *Manhua, Xudong et al's fingerprint search based on database clustering* [15]: Here, an efficient fingerprint search algorithm based on database clustering has been proposed. It non-uniformly partitions the fingerprint image by a circular tessellation to compute a multi-scale orientation field as the main search feature and average ridge distance used as auxiliary feature. Then it applies a modified *k*-means clustering technique to partition the orientation feature space into clusters. Based on the database clustering, they present a hierarchical query processing technique to facilitate an efficient fingerprint search. It speeds up the search process and also improves the retrieval accuracy.

[k] *Ji, Zhang Yi's SVM-based approach* [16]: Here, a classification method of fingerprint using orientation field and support vector machines is reported. It estimates the orientation field through pixel gradient values, and then calculates the percentages of the directional block of each class. These percentages are combined as a four dimensional vector, by which the trained hierarchical classifier that classifies the fingerprints into one of the six classes. The supervised training rule is adopted to train the hierarchical classifier with five one-against-many SVMs. Fingerprints can be classified into five classes by using the trained classifier.

[l] *Wei, Yonghui, et al's approach* [17]: This classification approach uses some curve features of the ridgelines. The approach basically exploits the total direction changes of the ridgelines during classification. However, the sampling of the ridgelines still can be found to be time consuming.

*Table 1* reports a general comparison of these classification techniques, discussed above.

Table 1: A general comparison of the fingerprint classification methods

| Ref. | Feature | Classes | Method (s) |
|------|---------|---------|------------|
| [6] | PCA* | 5 | ANN* |
| [7] | Singular point | 6 | Feature Based |
| [8] | Singular point | 4 | Feature Based |
| [9] | FingerCode | 5 | KNN & NN* Based |
| [10] | Singular point | 4 | Core Based |
| [11] | Coding matrix | 5 | ML* Based |
| [12] | Orientation field | 5 | GP* based |
| [13] | Orientation field | 5 | FbLD |
| [14] | DA* | 5 | FFT* and NDA* |
| [16] | Orientation field | 5 | SVM* |
| [17] | Curve features | 5 | Singularities |

*- PCA(Principal Component Analysis), ANN(Artificial Neural Network), GP(Genetic Programming), NDA(Non-linear discriminant analysis), FbLD (Feedback-based Line Detector), NN (Neural Network), ML (Machine Learning), DA(Discriminant Analysis), and SVM(Support Vector Machine).

Based on our limited survey, following observations are made:

(i)   Most of the existing methods classify the images based on the ridges, local features (i.e. minutiae) and global features (i.e. singular points).

(ii)  Model based approaches based on the global features (i.e. singular points) have been found more effective in classifying the fingerprints.

(iii) Structure-based approaches based on the estimated orientation field can be found capable to classify the enhanced images into one of the five classes.

(iv)  Cluster based search methods can be found to be effective in searching over large databases.

## 3. Background of the work

Real-time image quality assessment can greatly improve the accuracy of identification system. The good quality images require minor pre-processing and enhancement. Conversely, low quality images require major pre-processing and enhancement.

### 3.1 Fingerprint Image Pre-processing

Chen *et al.* [18] used fingerprint quality indices in both the frequency domain and spatial domain for image enhancement, feature extraction and matching performance. L. Hong [19] proposed the enhancement of the fingerprint image using filtering techniques. Lim *et al.* [20] computed the local orientation certainty level using the ratio of the maximum and minimum eigen values of the gradient covariance matrix and the orientation quality using the orientation flow. The main steps involved in the pre-processing include: (*a*) enhancement (*b*) binarization (*c*) segmentation, and (*d*) thinning. Next, we describe each of these steps, in brief.

(*a*) *Image Enhancement*: We have pre-processed the input fingerprint image on both the spatial and frequency domain. In the spatial domain, histogram equalization technique was applied for better distribution of the pixel values over the image to enhance the perceptional information. In the frequency domain, the image was enhanced based on the adaptive Fast Fourier transform (FFT) by blocks ($32 \times 32$ pixels). It improved the ridges and overall appearance of the image, which was found useful for extraction of quality features and hence for classification.

(*b*) *Image Binarization*: In this step, an *8*-bit grey level fingerprint image was transformed into a 1-bit per pixel image with 1-value for ridge and 0-value for furrow or valley. In [21], an optimized approach can be found for binarization. However, in our work, we used an enhanced binarization method which was based on the adaptive binarization approach [22]. In this method, we transformed the pixel value to 1 if the value is larger than the mean intensity value of the current block ($16 \times 16$ pixels), otherwise it is set to 0 (i.e. zero).

(*c*) *Image Segmentation*: The objective of this step basically is to extract the region of interest (ROI) which contains the desired fingerprint impression. Fingerprint image segmentation highly influences the performance of fingerprint identification system. Here, we used the gradient based fingerprint segmentation approach [23] which gives the accurate segmentation results for the low quality images also.

(*d*) *Image Thinning*: This step aims to eliminate the redundant pixels of ridges till the ridges are of just one pixel wide. We used an iterative, parallel thinning algorithm [24] for thinning the binarized fingerprint image.

In each scan of the full fingerprint image, the algorithm marks down redundant pixels in each small image window (3×3 pixels) and finally removes all those marked pixels after several scans. Also, in our experimentation, each step still has large computation complexity, although it does not require the pixel by pixel movement like other thinning algorithms.

## 3.2 Fingerprint Feature Extraction and Numeric Meta-base Creation

Fingerprint image offers a rich source of information for classification and matching of fingerprints. However, for a given raw image, automatic extraction of quality features is an extremely difficult task, specially, when the acquisitions are noisy. The effectiveness of fingerprint classification systems depend on the extracted features from the fingerprint images. There are mainly two types of features [25] that are useful in fingerprint identification systems: (*i*) Local features such as ridges and furrows details (minutiae), which have different characteristics for different fingerprints, and (*ii*) Global feature or pattern configurations which form the special patterns of ridges and furrows in the central region of the fingerprints. The first type of features carries information about the individuality of the fingerprints, whereas the second type carries information about the class of fingerprints. For effective recognition, the extracted features should be invariant to the translation and rotation of the fingerprint images. Mostly, global features can be derived based on the orientation and shape of the ridges. The orientation field of a fingerprint consists of the ridge tendencies in local neighborhoods and forms an abstraction of the local ridge structures.

### 3.2.1 Finding of Core or Reference Point

The singular points, i.e., *core* and *delta* points are unique structure of each fingerprint image, where the ridge curvature is higher than other areas and the orientation changes rapidly. The reference points are usually used for the classification purpose. However, some fingerprints, specially if the fingerprint image is partially captured then it is very difficult to locate the singular points. In the literature, researchers proposed many approaches for singular point detection and most of them are based on the fingerprint orientation field. The Poincare Index (PI) method [7] for singular point detection is one of the most commonly used techniques.

Based on our analysis on singular point detection techniques, it has been observed that the gradient based approach for the segmented images is efficient for orientation estimation (OE) as well as singular point detection [23]. So, we used this method for both OE and singular point detection. It first calculates the orientation field estimation by averaging the squared gradients of each block of the image. Then it applies the adaptive smoothen technique based on the analysis of the orientation consistency of the image to remove the noise, such as scars, ridge breaks, and low gray value contrast, etc. The reference point defined as the point with maximum curvature on the convex ridge, which is located in the central area of the fingerprint image. We used the grad-*x*, grad-*y* and directionality of the ridges based on the technique [8] as parameters of our method. So, we determined the core point based on the higher curvature area over the image based on the directional flow of the ridge pattern and it is basically for the five classes of fingerprint images such as *tented arch, left loop, right loop, whorl* and *twin loop*. We have used this method because of its simplicity, correctness, and easy to implement. The binarized image, orientation image and the core point of the image are shown in Fig. 1 (*a*), (*b*), and (*c*).

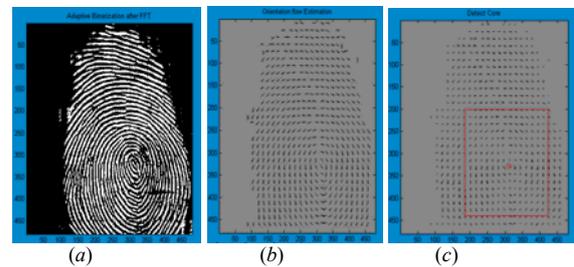

Fig. 1: (*a*) Binarized image, (*b*) Orientation image, and (*c*) Orientation image with reference or core point

### 3.2.2 Numeric Meta-base Creation

To create the meta-base, $T_{mxn}$, where *m* and *n* are the no. of rows (i.e. the number of images in the databases) and the number of dimensions respectively, we divided each pre-processed image into 3×3 grids (shown in Fig. 2(*a*)). Now, the initial grid cell i.e. $R_p$ is chosen which is basically the core or reference point in case of those five classes other than arch; however, in case of arch, it is the starting point of that ridge for which the inter-ridge distance can be found to be least. Then following this initial grid cell, the successive control points were chosen along the ridge flow of the image and accordingly numeric codes (0-7) are assigned depending on the regulation pattern and its directionality.

The idea behind the orientation flow codes is that each pixel of an image can have almost 8-neighbours and thus the direction from a given pixel can be specified by a unique number between 0 to 7 [2] (see Fig. 2(a)). Each of these numbers represents one of the eight possible directions from the given pixel along the ridge flow of the given image, and Fig. 2(*b*) & (*c*) shows the orientation image with and without grid space. This is continued for *n*-successive points. The parameters used in generating *T* and the corresponding values of the ridges (for the successive control points) and directionality for each numeric code are reported in *Table 2*. In our experimental study, we found better result for the number of control points i.e. $n = 32$ for representing an image.

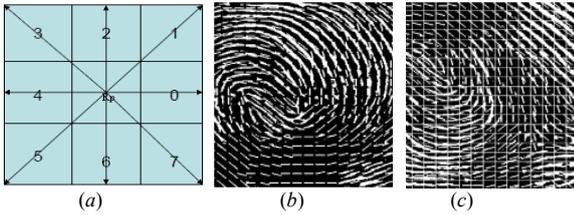

Fig. 2: (*a*) 3×3 grid and 8-connected codes, (*b*) Orientation flow image, and (*c*) Orientation flow image with 3×3 grid

An example of ridge flow pattern matrix is shown in Fig. 3. The logic for creation of $T_{mxn}$ is shown in procedure NMC. It takes image $I_i$ as input and invokes the procedure *GetCore* to find the core point over $I_i$. *GetCore* works based on [22]. It uses the variable $RPV_i$ to represent the ridge pixel value of the image and $Z_i$ to refer the condition instance of $Z$ values reported in *Table 2* so far. The variable $C_p$ represents the control points.

Table 2: Parameters for code generation and numeric code assignment

| Grad of x- $\delta_x$ | Grad of y- $\delta_y$ | Directionality, $Z_i = (4* (\delta_x + 2)+ (\delta_y + 2))$ | Ridge Values, $RPV_i$ | Numeric code |
|---|---|---|---|---|
| 0 | 1 | 11 | 1 | 0 |
| -1 | 1 | 7 | 1 | 1 |
| -1 | 0 | 6 | 1 | 2 |
| -1 | -1 | 5 | 1 | 3 |
| 0 | -1 | 9 | 1 | 4 |
| 1 | -1 | 13 | 1 | 5 |
| 1 | 0 | 14 | 1 | 6 |
| 1 | 1 | 15 | 1 | 7 |

$$T_{mxn} = \begin{matrix} 0\ 0\ 1\ 2\ 3\ 3.........6\ 7\ 7 \\ 0\ 4\ 5\ 5\ 7\ 7.........4\ 3\ 2 \\ 0\ 4\ 4\ 5\ 7\ 6.........3\ 3\ 2 \\ . \\ . \\ . \\ 0\ 0\ 1\ 1\ 3\ 4.........5\ 6\ 6 \end{matrix}$$

Fig. 3: Ridge flow pattern matrix

The procedure NMC is given below:

**Procedure *NMC*( )**
**Input**: Fingerprint image $I_i$ (i = 1, 2, 3,.......,m)
**Output**: Ridge orientation flow codes matrix, $T_{mxn}$
  1.   For $i = 1$ to $m$ do
  2.   Read the image $I_i$;
  3.   Call *GetCore*() to find the core point $C_i(x, y)$ ;
  4.   For each core point $C_i(x, y)$ do
     4.1 For $C_p = 1$ to $n$ do
       [a] if $RPV_i = 1$ and satisfies $Z_i$ then

          update $T_{mxn}$ with corresponding
          code (lookup *Table 1*)
     4.2 Next $C_p$;
  5.   Next *i*.

Next, we report our minutiae extraction and false minutiae removal technique.

## 4. FPMINU: Minutiae Extraction and False Minutiae Removal

The proposed FPMINU is an enhanced version of binarization based method [26] to extract the minutiae features from the fingerprint image using single pass. Initially, it converts the image into skeletonized (i.e. thinned) image. Then it takes one 5×5 window (unit pixel) and moves over the skeletonized image and count the number of unit pixels present in the subsequent ridges. We have chosen 5×5 window because of the following reasons:
(*i*) 3×3 window leads to confusion while identifying the bifurcation minutiae and triangle or delta (can be seen in *Fig. 4*)
(*ii*) 7×7 is too large and additional overhead occurs during true minutiae detection and false minutiae removal.

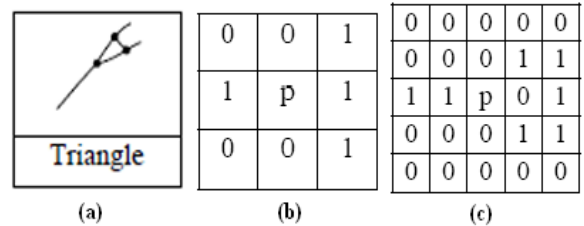

Fig. 4: False minutiae removal (*a*) Triangle (*b*) Using 3×3 window (*c*) Using 5×5 window

Over the 5×5 window, if the no. of unit pixels is 2 then it is identified as termination; on the other hand, if the no. is 6 then it is bifurcation; otherwise, it can be considered as false minutiae. For identification of the true minutiae, the following algorithm is used. This extracted minutiae features were used later for the local search within a cluster during identification.

**Procedure FPMINU()**
**Input**: Binarized thinned fingerprint image $I_i$
**Output**: Minutiae store in $M_{m \times n}$
1. for i = 1 to m do
2. read image $I_i$
3. Take a 5×5 window $W_i$ and move over $I_i$
4. Count no. of unit pixels $P_i$ on $W_i$;
    4.1 if $P_i < 3$ then
        store termination type $T_i$ in $M_{m \times n}$
    4.2 elseif $P_i < 7$ then
        store bifurcation type $B_i$ in $M_{m \times n}$
    4.3 else

store as false minutiae in $F_{m \times n}$
5. check the unit pixel $P_i$ positions for minutiae validation;
6. next i.

False minutiae may be introduced into the image due to several factors such as presence of noise due to thinning process, etc. Hence, after the minutiae are extracted, it is necessary to employ a post-processing stage in order to validate the minutiae. The experimental analysis of FPMINU in terms of the number of minutiae extracted or detected is shown in Fig. 5 (*a*), (*b*) & (*c*).

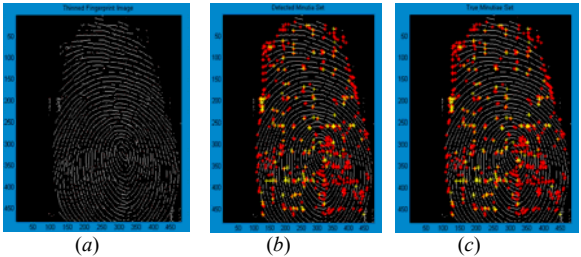

<div align="center">(<i>a</i>)        (<i>b</i>)        (<i>c</i>)</div>

Fig. 5: (*a*) Thinned image, (*b*) Detected minutiae set, and (*c*) True minutiae set

## 4.1 Effectiveness of FPMINU

To evaluate the performance of the proposed FPMINU, we computed minutiae over FVC datasets and due to lack of space a partial result is reported in *Table 3*. Except the first, our approach can extract larger no. of minutiae points than [27].

Table 3: A comparison of the recent fingerprint minutiae extraction method with FPMINU

| Fingerprint image | Gour et. al. [27] | FPMINU |
|---|---|---|
| 1_1.tif | 36 | 32 |
| 2_1.tif | 34 | 36 |
| 3_1.tif | 36 | 36 |
| 4_1.tif | 28 | 29 |
| 5_1.tif | 43 | 46 |

Next, we report an experimental study on unsupervised clustering based fingerprint classification method

## 5. Unsupervised Classification Approach

In this section, we report experimentation results over standard fingerprint datasets based on few popular clustering technique i.e. *K*-means, modified *K*-means, hierarchical clustering, DROCK and Fuzzy *C*-means. Standard *K*-means was tested for various dissimilarity measures and the best result of the technique has been shown in *Table 4*. Similarly, standard hierarchical technique (available with MATLAB) was tested for the various linkage measures and the result with the best one (i.e. single linkage) is reported in *Table 4*.

Table 4: A general comparison on various clustering techniques

| Dataset Size | K-means | Modified K-means | Hierarchical Clustering | DROCK |
|---|---|---|---|---|
| 32 | 0.125 | 0.093 | 0.125 | 0.023 |
| 300 | 0.060 | 0.033 | 0.186 | 0.038 |
| 800 | 0.150 | 0.070 | 0.087 | 0.041 |
| 1710 | 0.260 | 0.031 | 0.127 | 0.029 |
| 2300 | 0.204 | 0.022 | 0.120 | 0.047 |
| 2900 | 0.031 | 0.017 | 0.115 | 0.031 |

### 5.1 FPROCK: Proposed method

FPROCK classifies the images using an unsupervised approach i.e. DROCK [3] an agglomerative clustering technique, an enhanced version of ROCK [33]. FPROCK classifies the images based on the concepts of *neighbours* between the objects. Following definitions are relevant to our classification and searching approach.

*Definition* 1: Ridge flow pattern (RFP)
A RFP $Tt_{ij}$ of a ridge flow pattern matrix $T$ can be defined as an *n*-dimensional tuple where the value of each element $0 \le Lt_i \ge 7$;

*Definition* 2: Direct Neighbour of a RFP
A pair of pattern tuples $(Tt_i, Tt_j)$ can be defined as direct neighbours, if $sim(Tt_i, Tt_j) \ge \theta$; where $\theta$ is a user-defined threshold, and *sim* [ 3] can be defined as follows.

$$ sim = \begin{cases} 0, \ dissimilar \\ 1, \ identical \\ > 0, \ but \ < 1, \ similar \end{cases} $$

*Definition* 3: Indirect Neighbour
Let $Tt_i, Tt_j, Tt_k$ be three RFPs $\in D$, where $D$ is the metabase. Now, if $Tt_i$ is a direct neighbour of $Tt_j$ and $Tt_j$ is a direct neigbour of $Tt_k$, then $Tt_i$ is also neighbour (indirect) of $Tt_k$.

*Definition* 4: Candidate RFP
Any input RFP is defined as a candidate RFP, submitted for similarity search.

*Definition* 5: Prototype RFP
A stored RFP is defined as the prototype RFP used during similarity search.

*Definition* 6: Noise or Outlier
A RFP $Tt_i$ can be defined as a noise or outlier if $Tt_i \notin C$, where $C$ is the set of clusters i.e. $C = \{C_1, C_2, ...., C_6\}$ or , a RFP $Tt_i$ can be defined as noise if for any $Tt_j \in D$, $sim(Tt_i, Tt_j) < \theta$.

*Definition* 7: RFP Class or Cluster
A RFP class or cluster, $C_i$ can be defined as a set of RFPs where for any pair of RFPs say $(Tt_i, Tt_j)$, direct or indirect neighbour conditions (of *Definition 2 & 3*) are true.

*Definition* 8: RFP Profile of a Class
RFP profile $RP_i$ can be defined as a representation of the ridge flow pattern, minutiae representation, and core representation of the fingerprint image.
For any pair of candidate pattern tuples $p_i$ and $p_j$, $nhbr(p_i, p_j)$ is the linear relationship between them, where *nhbr* is the *neighbourhood* measure between a pair of objects. If this is close to 1 then both of them belong to the same cluster. If this is close to -1 then they are not similar. If this is equal to zero then they are not at all similar. Next, we report the FPROCK to classify the fingerprint images, in brief. This routine takes binary encoded $T_{mxn}$ i.e. $T'$ as input and generates *six* (here, k = 6) distinct classes i.e. $C_1$, $C_2$,.....$C_6$. Steps of the algorithm are:

**Procedure FPROCK( )**
*Input*   : encoded ridge flow data matrix $T'$
*Output* : classes $C_1$, $C_2$,...........,$C_6$
1. Read $T'$
2. *nhbr* := *compute_nhbr*($T'$);
3. for each s ∈ S do
    q[s] := *build_local_heap*(*nhbr*, s);
4. Q := *build_global_heap*(S, q);
5. while *size*(Q) > k do
    (*a*) u := *extract_max*(Q);
    (*b*) v := *max*(q[u]);
    (*c*) *delete*(Q, v);
    (*d*) w := *merge*(u, v);
    (*e*) for each x ∈ q[u] ∪ q[v] do
        (i) *nhbr*[x, w] := *nhbr*[x, u] + *nhbr*[x, v];
        (ii) *delete*(q[x], u); *delete*(q[x], v);
        (iii) *insert*(q[x], w, g(x, w)); *insert*(q[w], x, g(x, w));
        (iv) *update*(Q, x, q[x]);
    (*f*) *insert*(Q, w, q[w]);
    (*g*) *deallocate*(q[u]); *deallocate*(q(v));
6. end

**5.1.1 Complexity Analysis of FPROCK**
Considering the task of meta-base creation as a pre-processing operation, the complexity of our proposed classification method will be almost same with the DROCK or ROCK i.e. $O(n(llogn + n))$.

**5.1.2 Experimental Results**
(*a*) *Environment used*: The experiment was carried out on an hp xw8600 workstation with Intel Xeon Processor (*3.33*GHz) with 4GB of RAM. We used MATLAB *7.6* (R2008a) revised version in windows (*64*-bits) platform for the performance evaluation.
(*b*) *Datasets used*: Our algorithm was tested on the standard datasets: FVC 2000 [28], FVC 2002[29] and FVC 2004[30]. These datasets totally consist of *9600* fingerprint images from *330* different fingers, composed of various classes. However, we have chosen only *3000* images for experimentation. Out of these, *400* images are *arch*, *200* are *tented arch*, *890* are *left loop*, *900* are *right loop*, *510* are *whorl* and *100* are *twin loop*.
(*c*) *Results*: We have experimented with seven linkage measures such as *single, complete, average, weighted, centroid, median,* and *ward* with FPROCK over the FVC datasets and results are reported in *Table 5.1 & 5.2*. To evaluate the performance of the FPROCK method, we used the *misclassification error ($M_E$)* equation from [31].

$$M_E = \frac{1}{N}\sum_{i=1}^{k}\left|\left(cluster(D_i) - (cluster(D_i'))\right)\right|$$

where $D_i$ is the total number of cluster objects before clustering, $D_i'$ is the total number of cluster objects after clustering, $k$ is the total number of clusters and $N$ is the total number of records in the datasets.

It can be observed from the table that the performance of the FPROCK method for classification based on *complete* linkage is better than the other linkage measures.

Table 5.1:  Fingerprint classification error analysis using FPROCK

| Dataset Size | Single | Complete | Average | Weighted |
|---|---|---|---|---|
| 32 | 0.125 | 0.116 | 0.125 | 0.093 |
| 300 | 0.197 | 0.043 | 0.140 | 0.140 |
| 800 | 0.187 | 0.040 | 0.152 | 0.088 |
| 1710 | 0.190 | 0.059 | 0.149 | 0.130 |
| 2300 | 0.200 | 0.038 | 0.155 | 0.136 |
| 3000 | 0.187 | 0.027 | 0.140 | 0.140 |

**5.1.3 Clustering Effectiveness**

To evaluate the performance of FPROCK while comparing with its other counterparts, we determined accuracy and false acceptance rate (FAR) for several datasets of various sizes. The average result of the proposed method is reported in *Table 6*. From the table it can be seen that some of the methods [7],[9],[11],[30],[32] can be found to perform better, however, they can identify successfully four classes or five classes only. However, our method can identify six classes successfully at higher level of accuracy.

Table 5.2:  Fingerprint classification error analysis using FPROCK algorithm

| Dataset Size | Centroid | Median | Ward |
|---|---|---|---|
| 32 | 0.187 | 0.125 | 0.125 |
| 300 | 0.160 | 0.193 | 0.066 |
| 800 | 0.196 | 0.184 | 0.087 |
| 1710 | 0.187 | 0.173 | 0.093 |
| 2300 | 0.177 | 0.189 | 0.071 |
| 3000 | 0.160 | 0.193 | 0.069 |

An exhaustive analysis of the proposed method in terms of false acceptance rate (FAR) vs classification accuracy was also carried out in light of FVC datasets. The results and its comparison with its other counterparts are reported in Fig. 6. Next, we report our cluster based fingerprint search method.

Table 6: Comparisons results of FPROCK

| FPROCK and other counterparts | Classes | Accuracy(%) |
|---|---|---|
| Wilson et al. [32] | 5 | 81.0 |
| Karu et al. [7] | 5 | 85.4 |
| Jain et al. [32] | 5 | 90.0 |
| Hong et al. [9] | 5 | 87.5 |
| Cappelli et al. [32] | 5 | 92.2 |
| Yao et al. [11] | 5 | 89.3 |
| Chang et al. [32] | 5 | 94.8 |
| Mohamed et al. [32] | 5 | 92.4 |
| Zhang et al.[32] | 5 | 84.0 |
| Zhang et al.[32] | 5 | 84.0 |
| Yao et al.[32] | 5 | 90.0 |
| Wilson et al. [32] | 4 | 86.0 |
| Karu et al. [32] | 4 | 91.4 |
| Jain et al. [32] | 4 | 94.8 |
| Hong et al.[32] | 4 | 92.3 |
| Senoir [30] | 4 | 88.5 |
| Jain et al. [32] | 4 | 91.3 |
| Yao et al.[32] | 4 | 94.7 |
| Mehran et al. [32] | 4 | 99.02 |
| **FPROCK** | **6** | **97.3** |

## 6. Clustered Fingerprint Database Search

Based on our literature survey it has been observed that several novel attempts have been made in the past for effective search over large fingerprint databases using clustering approach. However, due to lake of the appropriate use of the proximity measures, such clustering based search methods have been found ineffective most of the time. In this paper, we proposed a fingerprint search method based on an agglomerative clustering approach (FPSEARCH).

### 6.1 FPSEARCH: Proposed Searching Method

The major purpose of using agglomerative hierarchical clustering is to facilitate the search over high dimensional numeric meta-space. Based on the formation of hierarchy of clustering using following approach, a hierarchy of RFP profiles is generalized. The conceptual view of the search strategy is shown in *Fig. 7*. Once query image is substantiated and relevant features of the image are extracted, the search process is initiated. The proposed search process works in two phases. In *phase-I*, a global search is conducted to identify the class of the queried image; then in *phase-II*, over a relevant cluster represented by the selected profile, a local search is carried out for the final verification.

Next, we report our two phase searching approach for fingerprint database. The first phase exploits the modified DROCK algorithm for clustering the database and hence the search space. Second phase deals with the low level searching over the clustered space using local features.

The fingerprint search is globally performed by comparing the query fingerprint i.e. the candidate RFP against all the RFP profiles by selecting the closest RFP profile.

### 6.1.1 *Phase-I*: Global Search

The global search is initiated with matching of the candidate RFP (Ridge orientation flow code generated for the queried image) against the prototype RFP profiles. It will identify the most relevant RFP profiles based on the number of agreements between candidate RFP and prototype RFP profile.

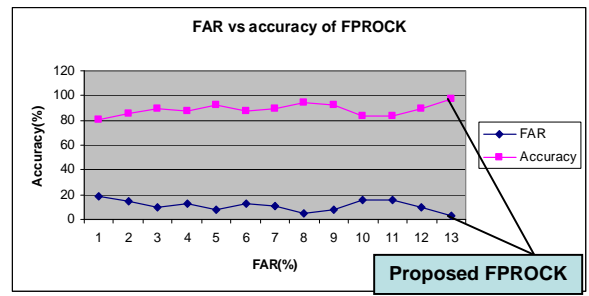

Fig. 6: FAR vs accuracy of FPROCK

### 6.1.2 *Phase-II*: Local Search Based on FPMINU

Local search based on true minutiae is initiated in the second phase over the clustered fingerprint search space for the individual verification of the queried fingerprint image. For each queried fingerprint image to initiate local search, it requires four parameters (α, β, γ, δ) where α is the number of true minutiae, β represents the total recurrence ridges, γ is the ROFC (ridge orientation flow code) and δ represents the core point of the image. Basically, the core point can be represented by dividing the whole image based on the some bit values [34] for each quadrant (we have taken 10 bits for our experiment) shown in *Fig 8* (for the example image, the bit pattern is 0110101010). Then, each candidate tuple i.e ($\alpha_c$, $\beta_c$, $\gamma_c$, $\delta_c$) will be matched against the stored prototype tuple ($\alpha_{ij}$, $\beta_{ij}$, $\gamma_{ij}$, $\delta_{ij}$) based on Euclidean distance where $i$ corresponds to the sub-cluster belonging to the respective class identified during *phase-I* (i.e. global search) and $j$ varies from 1 to $m$, where $m$ is the cardinality of the respective sub-cluster.

### 6.1.3 Effectiveness of FPSEARCH

The effectiveness of the clustering based fingerprint search approach, FPSEARCH was evaluated based on retrieval accuracy, penetration rate, and search complexity. The penetration rate is the average portion of the database retrieved over all query fingerprints. It indicates that how much the fingerprint search can narrow down the search space. In our approach, the query fingerprint was matched against each of the prototype profiles of the sub-cluster. If it doesn't match with an intermediate root node then the downward descendent or child nodes are avoided from further matching that way, it saves the search time significantly. The effectiveness of the searching in terms of penetration rate, number of comparisons, and retrieval accuracy over the clustered fingerprint database of various sample 300, 800, 1710, 2300, and 3000 based on the FVC datasets are reported in *Fig. 9 & 10*.

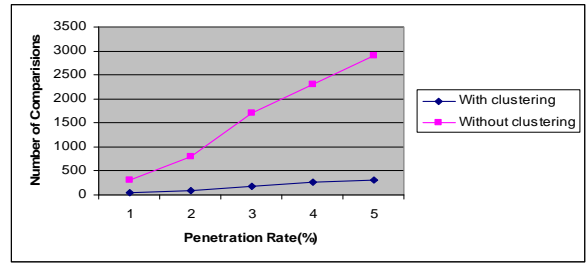

Fig. 9: Search Complexity: Fingerprint search (with clustering) vs Fingerprint search (without clustering)

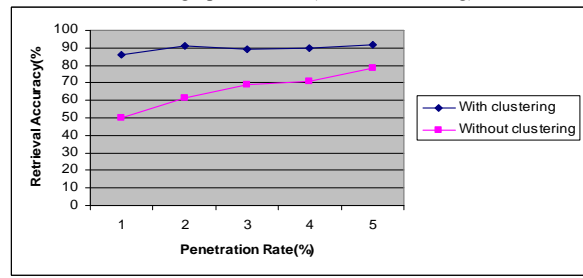

Fig. 10: Retrieval Accuracy: Fingerprint search (with clustering) vs Fingerprint search (without clustering)

## 7. Concluding Remarks

A limited survey on some of the popular fingerprint classification methods was carried out and reported in this paper. Most of these methods have been found capable of identifying four or five classes with an accuracy level of (80-95) %. This paper presents a fingerprint classification method with an accuracy level of 97%. Also, an effective search method over the clustered fingerprint database with high retrieval accuracy has been reported in this paper.

**Acknowledgment**

This work is a part of a research project and authors are thankful to *AICTE* for funding the project.

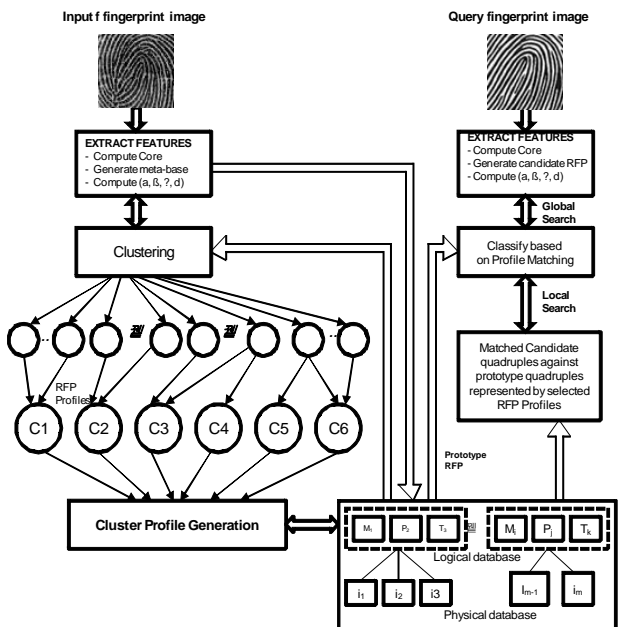

Fig. 7: Query based search over clustered fingerprint database

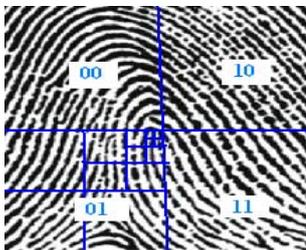

Fig. 8: Representation of Core point of the image

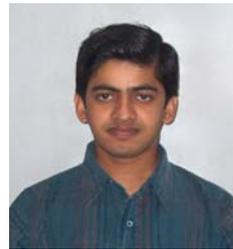

**Monowar Hussain Bhuyan** received his AMIETE degree in Computer Science & Engineering from The Institution of Electronics and Telecommunication Engineers (IETE) in 2007 and received his M.Tech. degree in Information Technology from Department of Computer Science & Engineering, Tezpur University in 2009. Now he is pursuing his Ph.D degree in Computer Science & Engineering from the same University. He is a life member of IETE, India. His research areas include Biometric Authentication, Data Mining, and Network Security.

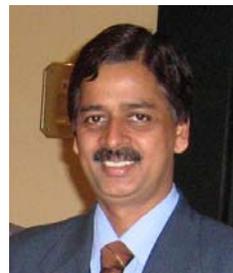

**Dr. Dhruba Kr Bhattacharyya** did his PhD in Computer Science from Tezpur University in 1999. Presently he is serving as a Professor in the Computer Science & Engineering Department at Tezpur University. His research areas include Data Mining, Network Security and Content based Image Retrieval. Prof. Bhattacharyya has published more than 100 research papers in the leading Int'nl Journals and Conference Proceedings. Also, Dr Bhattacharyya written/edited 04 books. He is a Programme Committee/Advisory Body member of several Int'nl Conferences/Workshops.